\documentclass[letterpaper, 10 pt, conference]{ieeeconf}  % Comment this line out if you need a4paper

\IEEEoverridecommandlockouts                              % This command is only needed if 
\usepackage[backend=bibtex,
            hyperref=true,
            url=false,
            isbn=false,
            doi=false,
            backref=false,
            style=ieee,
            natbib=true,%compatibility aliases
            mincitenames=1,
            maxcitenames=1,
            citestyle=numeric-comp,
            sorting=none,%nyt
            block=none]{biblatex}
        
\renewcommand{\bibfont}{\small}
\usepackage{flushend}
\usepackage{balance}

\usepackage{graphics} % for pdf, bitmapped graphics files
\usepackage{epsfig} % for postscript graphics files
\usepackage{amsmath} % assumes amsmath package installed
\usepackage{amssymb}  % assumes amsmath package installed

\usepackage{algorithm}
\usepackage{algorithmicx, algpseudocode}
\usepackage{color}
\usepackage{soul} %text highlighting
\let\labelindent\relax
\usepackage{enumitem}
\usepackage{xspace}

\newcommand{\eg}{\emph{e.g., }}

\newcommand{\figref}[1]{Figure\,\ref{fig:#1}}
\newcommand{\secref}[1]{Section\,\ref{sec:#1}}
\newcommand{\tabref}[1]{Table~\ref{tab:#1}}
\newcommand{\eqnref}[1]{{Eq.\ (\ref{eq:#1})}}

\newcommand{\pred}[1]{\texttt{\small{#1}}}

\DeclareMathOperator*{\argmax}{arg\,max}

\newcommand{\figtop}{\vspace*{-7mm}}
\newcommand{\figbot}{\vspace*{-2mm}}

\usepackage{etoolbox}
\makeatletter
\patchcmd{\@makecaption}
  {\scshape}
  {}
  {}
  {}
\makeatletter
\patchcmd{\@makecaption}
  {\\}
  {:\ }
  {}
  {}
\makeatother

\title{\LARGE \bf
Motion Reasoning for Goal-Based Imitation Learning
}

\author{% De-An Huang$^{1}$, \\XXX, and XXX$^{*,1,2,3}$% <-this % stops a space
De-An Huang$^{1,2}$, Yu-Wei Chao$^{*,2}$, Chris Paxton$^{*,2}$, Xinke Deng$^{2,3}$,\\
Li Fei-Fei$^{1}$, Juan Carlos Niebles$^{1}$, Animesh Garg$^{2,4}$, Dieter Fox$^{2,5}$
\thanks{
$^{*}$These authors contributed equally to the paper.
}
\thanks{
$^{1}$Stanford University, $^{2}$NVIDIA, $^{3}$UIUC, $^{4}$University of Toronto,
}
\thanks{$^{5}$University of Washington.}
}

\begin{document}

\maketitle
\thispagestyle{empty}
\pagestyle{empty}

%%%%%%%%%%%%%%%%%%%%%%%%%%%%%%%%%%%%%%%%%%%%%%%%%%%%%%
\begin{abstract}
We address goal-based imitation learning, where the aim is to output the symbolic goal from a third-person video demonstration. This enables the robot to plan for execution and  reproduce the same goal in a completely different environment. The key challenge is that the goal of a video demonstration is often ambiguous at the level of semantic actions. The human demonstrators might unintentionally achieve certain subgoals in the demonstrations with their actions. Our main contribution is to propose a motion reasoning framework that combines task and motion planning to disambiguate the true intention of the demonstrator in the video demonstration. This allows us to robustly recognize the goals that cannot be disambiguated by previous action-based approaches. We evaluate our approach by collecting a dataset of 96 video demonstrations in a mockup kitchen environment. We show that our motion reasoning plays an important role in recognizing the actual goal of the demonstrator and improves the success rate by over 20\%. We further show that by using the automatically inferred goal from the video demonstration, our robot is able to reproduce the same task in a real kitchen environment.  

\end{abstract}

%%%%%%%%%%%%%%%%%%%%%%%%%%%%%%%%%%%%%%%%%%%%%%%%%%%%%%%%
\section{Introduction}

We are interested in allowing robots to learn new tasks from video demonstrations.
Recently, there has been rapid progress in imitation learning~\cite{abbeel2004apprenticeship,ho2016generative,ziebart2008maximum,finn2016guided}, which even enables learning a new task from a \emph{single} demonstration of the task~\cite{duan2017one,finn2017one,xu2018neural}. By leveraging meta-learning~\cite{finn2017model}, the robot learns to follow the actions in the demonstration. In many cases, however, the robot does not have to thoroughly follow the actions in the demonstration to complete the task. Instead, what matters more is the \emph{goal} or the intention of the demonstrator~\cite{cheng2019purposive,chung2015bayesian,pathakICLR18zeroshot,verma2006goal}.
For example, if the goal is to get a bowl, it does not matter which hand we use to pick the bowl, and whether we get the bowl from the cabinet or the dishwasher. Understanding the intention of human demonstrator is important for human-robot interaction~\cite{dragan2013legibility} and can enable the robot to generalize a wider range of scenarios~\cite{cheng2019purposive}.

\begin{figure}[t]
  \centering
  \includegraphics[width=1.0\linewidth]{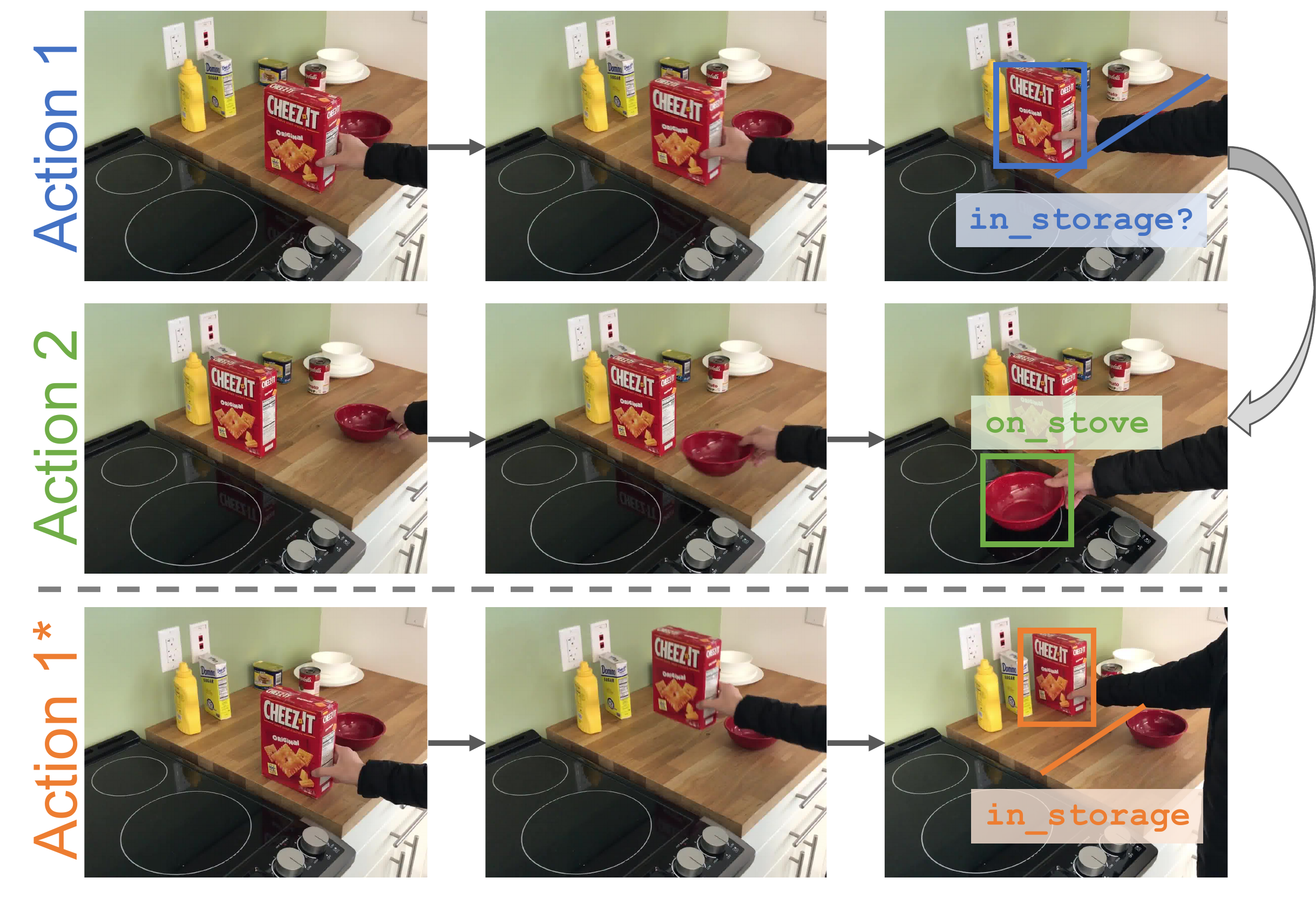}
  \figtop
  \caption{
    The first action moved the cracker box into the storage and the second action moved the bowl onto the stove. However, \pred{in\_storage} is not necessarily intentional. One can also explain that Action 1 is moving the cracker box out of the way for Action 2. This is even more obvious if we compare Action 1 with Action 1$^*$ in the last row.
  }
  \figbot
  \label{fig:fig1}
\end{figure}

There has been many works that aim to infer the intention of humans in both robotics~\cite{chung2015bayesian,katz2017novel,koppula2015anticipating,ramirez2009plan,ziebart2010modeling} and cognitive science~\cite{baker2009action,goodman2016pragmatic,solway2012goal}. However, these works are mostly limited to trajectory prediction in 2D environments, and goal recognition from real-world videos remains challenging.
The main challenge of applying goal reasoning to real-world human demonstration is that the goal or the intention can be ambiguous and cannot be fully determined by either or both the final state and the sequence of high-level actions. For 2D trajectory prediction, it is reasonable to assume that the final state defines the goal because the goal is just the final location of the object. However, this is not the case in real-world demonstrations. Take the kitchen environment in \figref{fig1} as an example. There are many objects (sugar box, bowl, etc) in the scene, and it is unclear what the goal of the demonstration is from just the final state or configuration.

An alternative is to recognize the sequence of actions performed in the demonstration~\cite{chung2015bayesian}. In \figref{fig1}, the person first moved the cracker box to the storage region with Action 1, and moved the bowl onto the stove with Action 2. Based on these actions, one might infer that the goal is: \pred{in\_storage(cracker\_box)}, \pred{on\_stove(bowl)}. However, the interpretation is not always unique. One can also interpret Action 1 as just moving the cracker box out of the way, so that it is not blocking Action 2 to move the bowl onto the stove. In this case, despite the person moved the cracker box, it is not part of the goal. This becomes more clear if we compare Action 1 with Action 1$^*$ in the last row. It is more likely in Action 1$^*$ that the person is intentionally moving the cracker box into the storage. We as humans can reliably interpret others' intentions despite the multiple explanations for their actions. This ability to reason about the actual intention in ambiguous scenarios allows us to interact smoothly and generalize to drastically different scenarios. 

The next question is: How can our system differentiate Action 1 and Action 1$^*$? Or how can our system  determine if the person is moving the cracker box intentionally to achieve \pred{in\_storage(cracker\_box)}  or just moving it out of the way in \figref{fig1}? Prior plan recognition approaches~\cite{ramirez2009plan,sohrabi2016plan,yolanda2015fast} are not directly applicable because the trajectory of the demonstration is already completed. On the other hand, action recognition based approaches~\cite{chung2015bayesian,katz2017novel}, which model the movements of the objects independently, are not capable of determining whether moving the cracker box into the storage is intentional or not.

Our key observation is that by moving the cracker box the demonstrator actually achieves \emph{two} things simultaneously. In addition to \pred{in\_storage(cracker\_box)}, the new location of the cracker box also makes it \emph{out of the way} from moving the bowl onto the stove. By providing this alternative explanation of the demonstrator's action, it becomes less likely that the demonstrator moves the cracker box to achieve \pred{in\_storage(cracker\_box)}. We refer to being \emph{out of the way} as a \emph{motion predicate} achieved by moving the cracker box to differentiate it from the standard predicates. We refer to the standard predicates like \pred{in\_storage(cracker\_box)} as \emph{task predicates}. Consequently, the question is to decide whether an action aims to achieve the motion predicate or the task predicate or both.

Despite the potential alternative interpretations of the demonstration, we assume the demonstrator aims to cooperatively and unambiguously communicate the task goal through the demonstration~\cite{grice1975logic}. In other words, the demonstrations are still \emph{intent-expressive} or \emph{legible}~\cite{dragan2013legibility}. However, as we have discussed, the intention cannot be fully captured by the high-level actions in cases like \figref{fig1}.

Our main contribution is to observe that the legibility of such demonstrations thus resides in the low-level trajectories or motions instead of the high-level actions. This legible motion hypothesis allows us to formulate the decision between task and motion predicates as inverse planning~\cite{baker2009action}.

We evaluate our hypothesis by collecting a dataset of real video demonstrations conditioned on a given set of goals within the kitchen domain. Our results show that our inverse planning formulation based on the task and motion predicates is able to reliably infer the intention or the goal of demonstrations. 
This provides the fundamental basis for goal-based imitation learning from real-world videos. To demonstrate its utility, we apply our goal recognition approach to address third-person imitation from observation~\cite{liu2018imitation,stadie2017third}, where the robots need to execute the tasks based on demonstrations from different environments and demonstrators.
Based on the demonstrations collected in a mockup tabletop kitchen, our robotic system is able to infer the demonstrated goal or high-level concepts~\cite{lazaro2018beyond} and reproduce the same goal with a robot in a real kitchen.

\section{Related Work}

\noindent \textbf{Goal and Intention Recognition.} We aim to recognize the goal or intention of a video demonstration. Related problems have been explored in plan and goal recognition~\cite{ramirez2009plan,sohrabi2016plan,sukthankar2014plan,yolanda2015fast,ziebart2008maximum,koppula2015anticipating}, goal-based imitation~\cite{verma2006goal,chung2015bayesian,cheng2019purposive}, and Bayesian Theory of Mind~\cite{baker2011bayesian,lee2018bayesian}. Understanding the intentions of the agents is important for multi-agent systems~\cite{lanctot2017unified,rabinowitz2018machine} and human-robot interaction~\cite{kelley2008understanding,ramachandran2016shaping}.
In our work, we introduce motion reasoning to task-based goal recognition. This allows us to go beyond early recognition of 2D trajectory end points.

\vspace{1mm}
\noindent \textbf{Imitation from Observation.}
We recognize the goal of a video demonstration and use it for the robot to imitate the demonstrated task. This imitation-from-observation~\cite{liu2018imitation,torabi2019adversarial} setup does not require the demonstrations to be from the same agent, and not even the same environment~\cite{stadie2017third,sermanet2016unsupervised}. Moreover, we address goal-based imitation learning from just a single demonstration of the task. This is related to recent progress on one-shot imitation learning~\cite{duan2017one,finn2017one,xu2018neural,huang2019neural}. Instead of collecting large amount of training data for end-to-end training, we explicitly reason about the object trajectories in the demonstration, which is more data-efficient.

\vspace{1mm}
\noindent \textbf{Interpretable Robot Motion.}
We resolve the goal ambiguity of high-level actions by reasoning about the low-level object trajectories, assuming the trajectories are intent-expressive or legible~\cite{chakraborti2019explicability}. Generating intent-expressive actions has been an important area of human-robot interaction~\cite{dragan2013legibility,szafir2014communication,tellex2014asking,zhang2017plan,macnally2018action} because the generated actions need to be unambiguously communicated to robots. Instead of generating these motions, we aim to \emph{recognize} intent-expressive trajectories. 

\vspace{1mm}
\noindent \textbf{Task and Motion Planning (TAMP).}
% Caelan
We introduce motion reasoning to task goal recognition. We treat the demonstrator as a task and motion planning (TAMP)~\cite{garrett2018sampling,toussaint2018differentiable,dornhege2013lazy} agent instead of just a motion~\cite{baker2009action} or task planning~\cite{katz2017novel} agent when recognizing the goal. Our motion predicate indicates how the poses of the objects affect the trajectories of other actions, which is related to the semantic attachments~\cite{dornhege2013lazy} and the geometric constraints~\cite{lozano2014constraint} in TAMP.

\section{Method}

\begin{figure*}[t]
  \centering
  \includegraphics[width=1.0\linewidth]{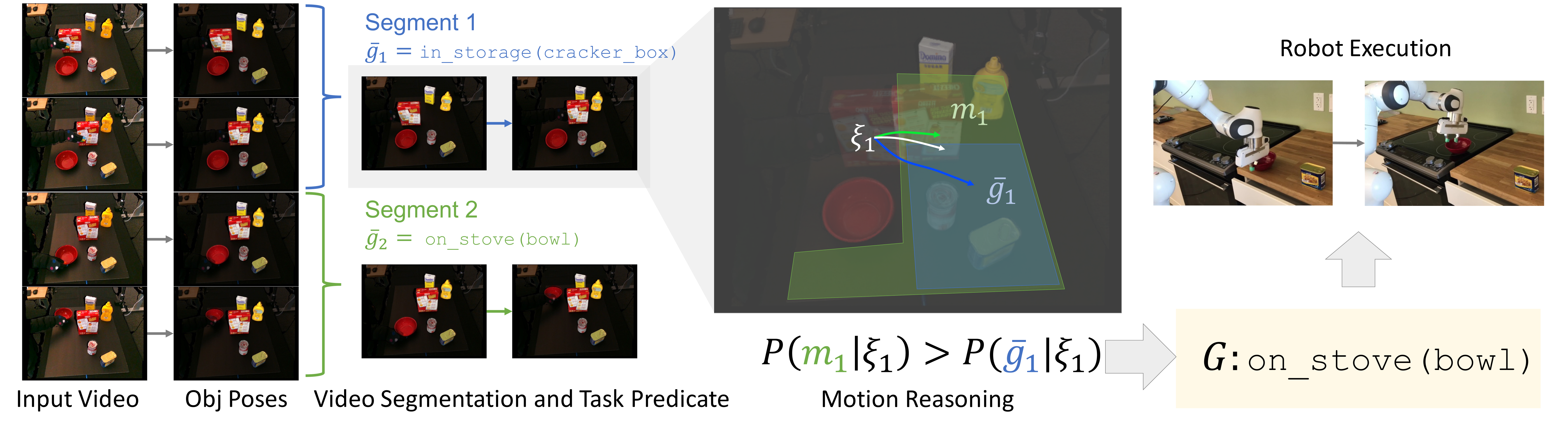}
  \figtop
  \caption{
    Overview of our approach. Given an input video demonstration, we first detect the object poses and temporally segment the video to recognize the task predicate $\bar{g}_i$ for each segment $S_i$. We then reason if the observed object trajectory $\xi_1$ can be better explained by the motion predicate $m_1$ (not blocking $\xi_2$) or the task predicate $\bar{g}_1$. Once we have all the intentional task predicates, we can pool them together to obtain the goal $G$ of the demonstration. This can be used for robot execution in a different environment.
  }
  \figbot
  \label{fig:system_fig}
\end{figure*}

We address goal-based imitation learning from real-world videos. Given a video demonstration of a task, we aim to output the symbolic goal of the task. This symbolic goal can then be used as input for robotics systems to reproduce the task in potentially different environments. The main challenge of goal recognition from real-world videos is that there exist  multiple interpretations of the goal based on just the final state and the high-level actions in the demonstration. Our key observation is that the low-level motion trajectories are thus intent-expressive. This allows us to formulate the problem as recognizing whether the motion predicate or the task predicate or both are generating the trajectories. By performing motion reasoning based on inverse planning, we are able to interpret the actual goal of the demonstration beyond just the final state and the sequence of high-level actions. \figref{system_fig} shows an overview of our approach.

We will first discuss our goal-based imitation from observation setup in \secref{goal_imit}. This goal-based formulation enables learning from video observations of different environments and contexts.
Next, we will introduce the proposed motion predicate for goal-based imitation learning and how we can use motion reasoning to determine if the task or the motion predicate is the intended subgoal for an action in \secref{motion_reasoning}. 
Finally, we will include details of our visual perception pipeline in \secref{percept}.

\subsection{Goal-Based Imitation from Observation}
\label{sec:goal_imit}

Given a demonstration $D = [z_1, \dots, z_T]$ of length $T$, the aim of goal-based imitation~\cite{verma2006goal} is to output the intended goal $G$ of the demonstration. We follow the imitation-from-observation setup~\cite{liu2018imitation}, and thus the elements $z_t$ in the demonstration do not necessarily correspond to the agent's state. In our setup, $z_t$ is a video frame of a third-person human video demonstration $D$. We assume that both the human demonstration and the robot execution are based on the same task planning domain~\cite{ghallab2004automated} (\eg share the same PDDL domain file). The domain contains a list of ground atoms or predicates (\eg \pred{in\_storage(cracker\_box)}, \pred{on\_stove(bowl)}) and a set of grounded operators.
Each grounded operator consists of its name, the list of arguments, the precondition of a list of predicates that need to be satisfied in order to apply the operator, and the effects of how the state would change by applying the operator. We also refer to the grounded operators as (high-level) actions.

Based on this definition, a goal $G = [g_1, \dots, g_N]$ consists of a list of $N$ predicates $g_i$ that need to be true to achieve the goal. We also refer to $g_i$ as \textit{subgoal} or \textit{task predicate}. We assume that $g_i$ can only be selected from a known subset of all the predicates in the domain. 

This goal-based formulation naturally enables imitation from observation in different environments. Given a video demonstration $D_{e_1}$ in environment $e_1$, as long as we can extract the goal $G$, then $G$ can be used to define the task planning problem in a new environment $e_2 \neq e_1$. The robot can then execute the output plan in $e_2$ based on the goal $G$ extracted from $D_{e_1}$. In addition, $G$ is also independent of the agent's state and motion as long as the task domain definition is shared between $e_1$ and $e_2$. In our experiment, the human demonstration can be in a remote mockup kitchen $e_1$, while the robot execution is done in the real kitchen $e_2$
This generalizability across environments and agents is a key feature of our approach.

\subsection{Motion Reasoning for Video Goal Recognition}
\label{sec:motion_reasoning}

We have discussed our goal-based imitation from observation setup, and how it can be used for execution in different environments based on third-person human demonstrations. Now we discuss how we go from the demonstration $D = [z_1, \dots, z_T]$ to the underlying goal $G$. Our approach consists of three main steps: 
(i) First, the demonstration $D$ is temporally segmented into video segments $\{S_i\}$. 
Without loss of generality, we assume in this section that each of the segment $i$ achieves one of the task predicates $\bar{g}_i$ by manipulating a single object.
(ii) Second, for each of the video segment, the manipulation of the object also achieves a motion predicate in addition to the task predicate. We then extend inverse planning to decide if the video segment is intended to achieve the task predicate $\bar{g}_i$ or the motion predicate or both. 
(iii) Finally, once we know all the intentional task predicates $\bar{g}_i$ achieved by all the segments $S_i$, we can pool them together based on the domain definition to get the final goal $G = \{g_i\}$. 
An overview of our approach is shown in \figref{system_fig}. 

For video human demonstration, step (i) is equivalent to action segmentation~\cite{kuehne2014language}, which is a well-developed field in computer vision~\cite{chang2019d3tw,Damen2018EPICKITCHENS}. We will directly go into our main contribution step (ii) in this section, and the details for step (i) will be discussed in \secref{percept}.

Assuming that we have temporally segmented the demonstration based on the changes of task predicates, we then have the segment $S_i=[z_1^i,\dots, z_t^i,\dots, z_{T_i}^i]$ consist of the observations $z_t^i$ and the task predicate $\bar{g}_i$ achieved in this segment. Based on the arguments of the task predicate, we also know the object $b_i$ being manipulated in this segment. Our next step is to decide if the task predicate $\bar{g}_i$ achieved in the segment $S_i$ is actually part of the goal of the demonstration. Previous works for goal recognition from real-world demonstration~\cite{chung2015bayesian,katz2017novel} focus on the relationships between different predicates $\bar{g}_i$. However, as we have discussed, for cases like \figref{fig1}, we are unable to determine the intention simply based on $\bar{g}_i$. Despite the ambiguity at the predicate level, we do know that the aim of a demonstration is to communicate the goal. In this case, there should exist other information to resolve this ambiguity at the predicate level. 

We propose that the solution is to reason at the motion or the object trajectory level between the segments. We observe that in addition to the task predicate $\bar{g}_i$, the manipulation of object $b_i$ also achieves something else. Now $b_i$ is at a new location and has a new pose. Independent of the task predicate $\bar{g}_i$, this new pose of $b_i$ can also enable object trajectories in later segments. Intuitively, we can think of high-level action based approaches~\cite{chung2015bayesian,katz2017novel} as just considering the task planning aspect of goal recognition, and 2D trajectory approaches~\cite{ramirez2009plan,ziebart2010modeling} as just considering the motion planning aspect of goal recognition. Our approach instead considers both the task and motion planning aspects to infer the goal of the demonstration. In this case, although moving the bowl onto the stove in the next segment $S_{i+1}$ might symbolically only includes \pred{in\_hand(bowl)} as its precondition, there is also implicitly a motion constraint~\cite{lozano2014constraint} of valid paths that enables the moving of the bowl onto the stove. We thus call satisfying the constraint by moving $b_i$ to create the valid path the \textit{motion predicate} $m_i(b_i)$ achieved by the segment $S_i$.

There are easy cases: If moving the object $b_i$ achieves a task predicate and does not create any valid paths for other objects then the intention is just the task predicate. On the other hand, if no new task predicate is achieved and new valid paths are created by moving $b_i$ then the intention is the motion predicate. The challenging case is: when both task predicate $\bar{g}_i$ and the motion predicate $m_i(b_i)$ becomes true in $S_i$, which one is the actual intended goal? Consider  \figref{fig1} again, moving the cracker box both achieved \pred{in\_storage(cracker\_box)} and create a valid path for moving the bowl onto the stove. How do we know if \pred{in\_storage(cracker\_box)} is part of the goal?

After explicitly formulating the motion predicate, we can now apply the principle of rational actions~\cite{csibra2007obsessed}: we can assess whether the motion predicate or the task predicate would be more efficiently brought about by the observed object manipulation trajectory. This is in line with Bayesian inverse planning~\cite{baker2009action}. Let 
$\xi^i_{s\rightarrow q}$ be the trajectory of $b_i$ in $S_i$ starting from $s$ and ending at $q$, we can decide the intention of $S_i$ by:
\begin{equation}
    \argmax_{g\in \{\bar{g}_i, m_i(b_i)\}} P(g|\xi^i_{s\rightarrow q}) = P(\xi^i_{s\rightarrow q}|g) P(g)
\end{equation}
Following~\cite{dragan2013legibility}, we can derive
\begin{equation}
\label{eq:goal_recog}
    P(\xi^i_{s\rightarrow q}|g) \propto \frac{\exp(-C(\xi^i_{s\rightarrow q}) -C(\xi^{i*}_{q\rightarrow g}))}{\exp(-C(\xi^{i*}_{s\rightarrow g}))},
\end{equation}
where $\xi^{i*}_{s\rightarrow g}$ and $\xi^{i*}_{q\rightarrow g}$ are the optimal trajectories to achieve $g$ from $s$ and $q$, and $C(\cdot)$ is the function to compute the cost of a trajectory. We obtain the object trajectories by tracking the pose of each object. Each frame $z^i_t$ is represented by the poses of all the objects in the scene (details in \secref{percept}).

\begin{figure}[t]
  \centering
  \includegraphics[width=1.0\linewidth]{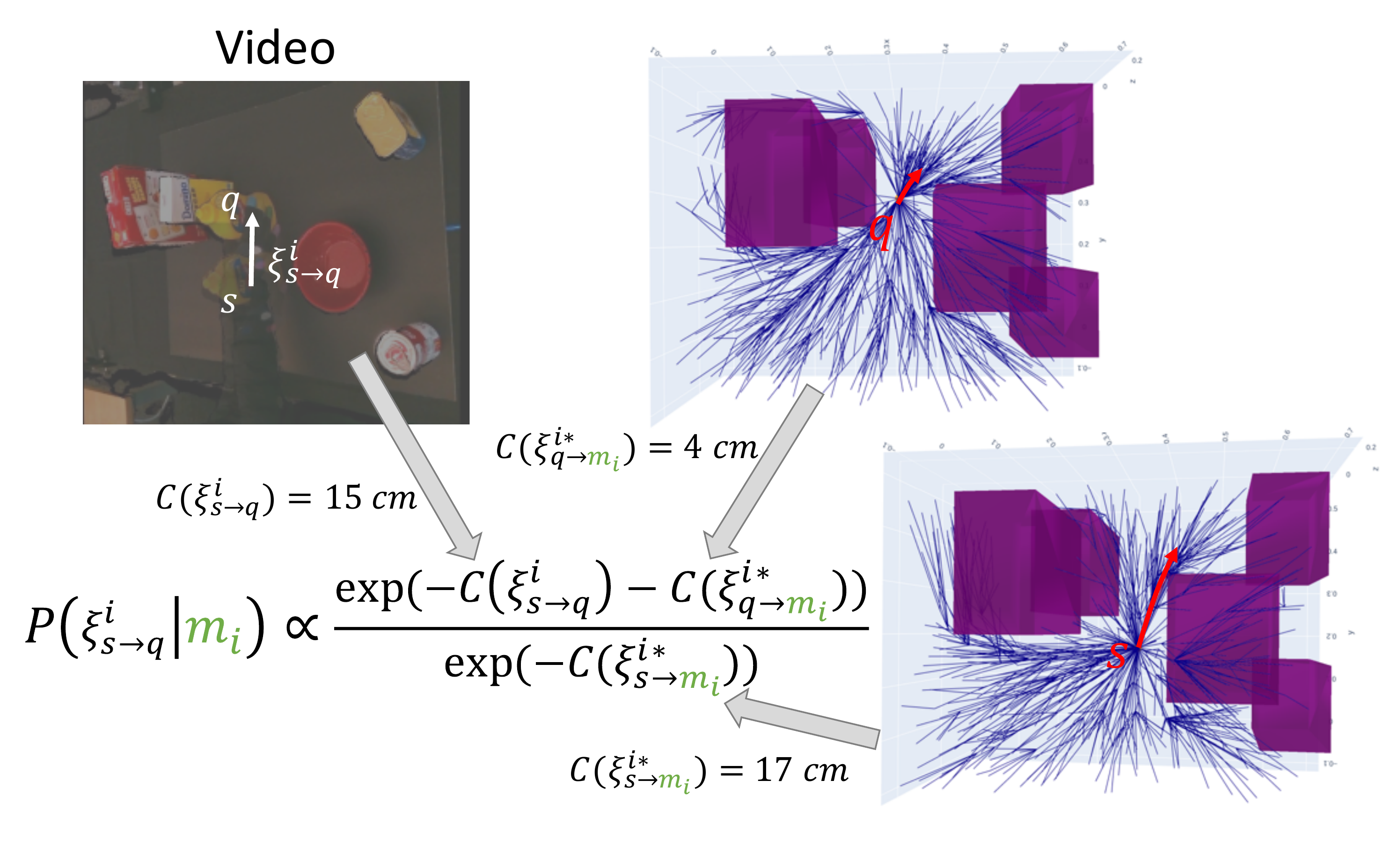}
  \figtop
  \caption{
    Example of our motion reasoning for $P(\xi_{s\rightarrow q}^i|m_i)$. The cost of the trajectory $C(\xi_{s\rightarrow q}^i)$ can be estimated from the video. We use RRT* to estimate the optimal costs $C(\xi^{i*}_{s\rightarrow m_i})$ and $C(\xi^{i*}_{q\rightarrow m_i})$. $P(\xi_{s\rightarrow q}^i|\bar(g)_i)$ can be computed similarly.
  }
  \figbot
  \label{fig:motion}
\end{figure}

When $g$ is just a location in space, $\xi^{i*}_{s\rightarrow g}$ and $\xi^{i*}_{q\rightarrow g}$ are more straightforward to compute. In our case, $g$ in \eqnref{goal_recog} is either the task predicate or the motion predicate, which can be satisfied by a region in space instead of a single point. In this case, it is inefficient to directly discretize the space and run search algorithms. We thus use RRT*~\cite{karaman2011sampling} from $s$ and $q$ to approximate the optimal trajectories to achieve $\bar{g}_i$, $m_i(b_i)$. In order to run RRT*, we first treat other objects $b_j$, $j \neq i$ as obstacles and use the object poses at frame $t$ to compute the configuration space. When $\bar{g}_i$ is a region in space, the success condition for RRT* is to reach the region. On the other hand, $m_i(b_i)$ means that $b_i$ is not blocking the trajectory of $b_{k}$, , $k \neq i$. We find the convex hull that covers the trajectory of $b_{k}$, and use RRT* to find the shortest path that $b_i$ is not intersecting with this convex hull. \figref{motion} shows an example that applies \eqnref{goal_recog} to compute $P(\xi^i_{s\rightarrow q}|m_i(b_i))$ 

\vspace{1mm}
\noindent\textbf{Task Predicate Pooling.} Using the proposed motion predicate reasoning, we are able to decide for each segment $S_i$ if the corresponding $\bar{g}_i$ is intentional. The final step is then pooling all the intentional $\bar{g}_i$ into the final goal $G$. We combine the intentional task predicates by removing the ones that are preconditions for later intentional task predicates. For example, although one moves the bowl onto the stove, the actual goal might just be to cook what is inside the bowl. Later the bowl is moved back on to the table to serve and being on stove is thus not part of the final goal.

\subsection{Visual Perception Pipeline}
\label{sec:percept}

We now discuss our visual perception pipeline and how we use its output to  segment the video.

\vspace{1mm}
\noindent\textbf{6D Pose Estimation.} As our approach reasons at the level of object trajectories, we need to first detect and track the object poses in the 3D space. We initialize the object poses with PoseCNN~\cite{xiang2017posecnn}. The detection output is then used as initialization for PoseRBPF~\cite{deng2019poserbpf} to track the 6D poses. The tracked poses are further optimized based on the signed distance functions~\cite{schmidt2014dart}. When multiple cameras are available, we use the maximum particle score~\cite{deng2019poserbpf} to select the best view for the object. In this case, we transform the video frame observation $z_t$ to an object-centric representation $x_t = \phi(z_{1:t}) = [\hat{x}_t^1, \dots,\hat{x}_t^k, \dots ,\hat{x}_t^K]$. Here $\phi$ is our object pose tracking pipeline, $x_t$ is the object-centric representation of $z_t$ and $\hat{x}_t^k$ is the estimated 6D pose of the $k$-th object. Note that one can further augment $x_t$ with detected hand trajectories~\cite{handtracking} to improve the robustness of the downstream video segmentation task.

\begin{figure*}[t]
  \centering
  \includegraphics[width=1.0\linewidth]{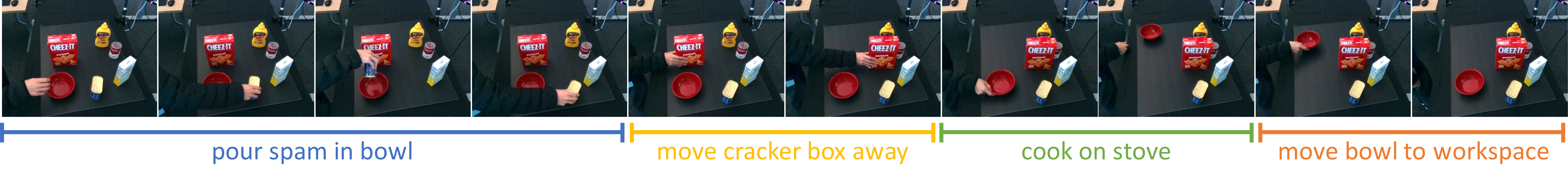}
  \figtop
  \caption{
    Example of our demonstration in the mockup tabletop kitchen. The person first pours the spam into the bowl and moves the cracker box away so that the bowl can be moved on the stove to cook the spam. Finally the bowl is moved back to the workspace.
  }
  \figbot
  \label{fig:demo}
\end{figure*}

\begin{figure}[t]
  \centering
  \includegraphics[width=1.0\linewidth]{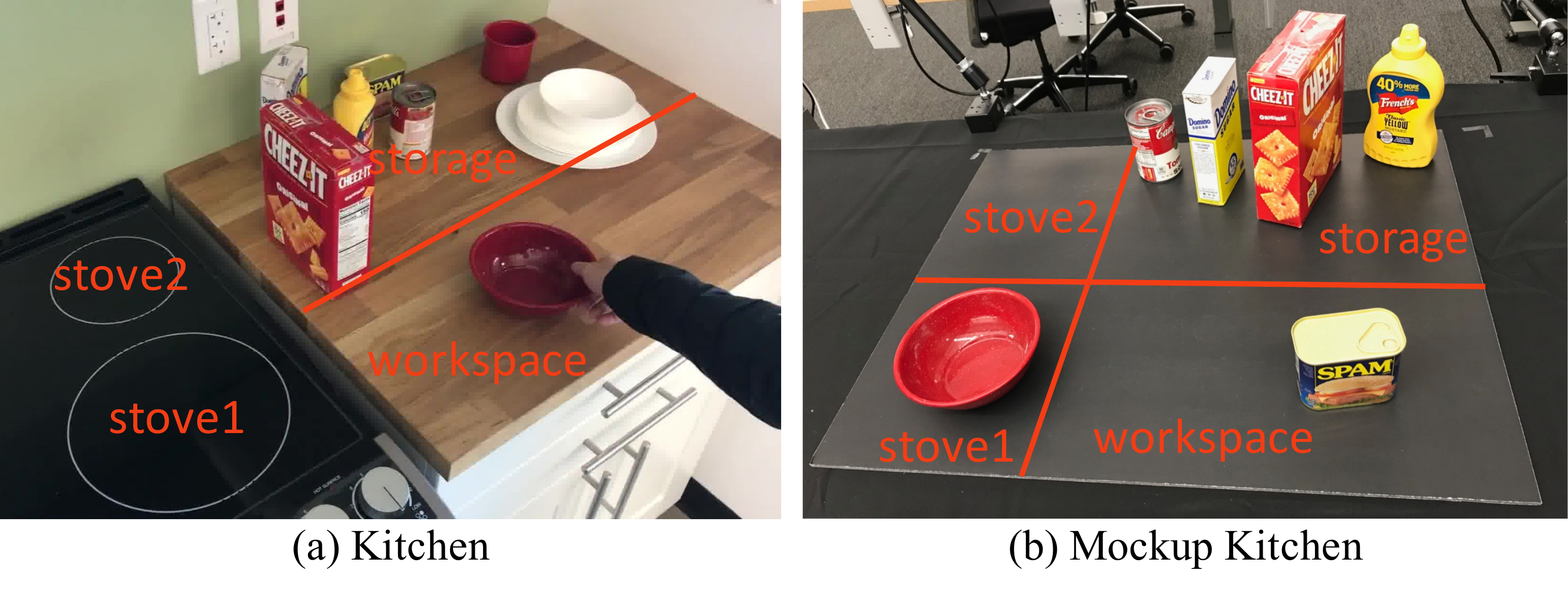}
  \figtop
  \caption{
    Our cooking domain involves four regions: the \emph{workspace} that is closest to the agent, the \emph{storage} that is further away, and two \emph{stoves} on which ingredients can be cooked.
  }
  \figbot
  \label{fig:setup}
\end{figure}

\vspace{1mm}
\noindent\textbf{Temporal Segmentation.} The first step of our approach is to temporally segment the demonstration so that we can reason about the trajectories between the segments in the following steps. While one can treat this as an action segmentation problem~\cite{kuehne2014language} and collect annotated data for training neural networks, we choose to segment the demonstration based on the object pose trajectories we extract for motion reasoning. For each time step $t$, we have the poses $\hat{x}_t^k$ for all the objects. By comparing $\hat{x}_t^k$ temporally, we are able to know if object $k$ is moving and being manipulated. We then segment the video such that each segment contains the manipulation of a single object. As earlier noted, the segmentation outcome can be further improved by refining the temporal boundaries based on the detected hand-object distances. Next, we compute the predicates for each frame based on the estimated poses.  By comparing the predicates at the start and the end of the segment, we can get the task predicate(s) for the segment.

\section{Experiments}

The aim of the paper is to recognize the goal of video demonstrations even when it is not obvious from the final state and the high-level actions. We hypothesize that the object trajectories are thus intent-expressive and propose a motion reasoning framework to recognize the goals of the demonstrations. Our experiments aim to answer the following questions: (1) Are there task predicates that are not intentional in the demonstrations? (2) Can our motion reasoning framework determine if a task predicate is intentional? (3) Can we address third-person imitation from video observation? We answer the first two questions by collecting a new dataset consist of demonstrations of a mockup cooking task. We then apply our motion reasoning framework to recognize the goals of the demonstrations and compare to existing approaches for goal recognition. We answer the last question by performing robot execution based on our extracted goal in a different environment.

\subsection{Mockup Cooking Domain}

\begin{figure*}[bt]
  \centering
  \includegraphics[width=1.0\linewidth]{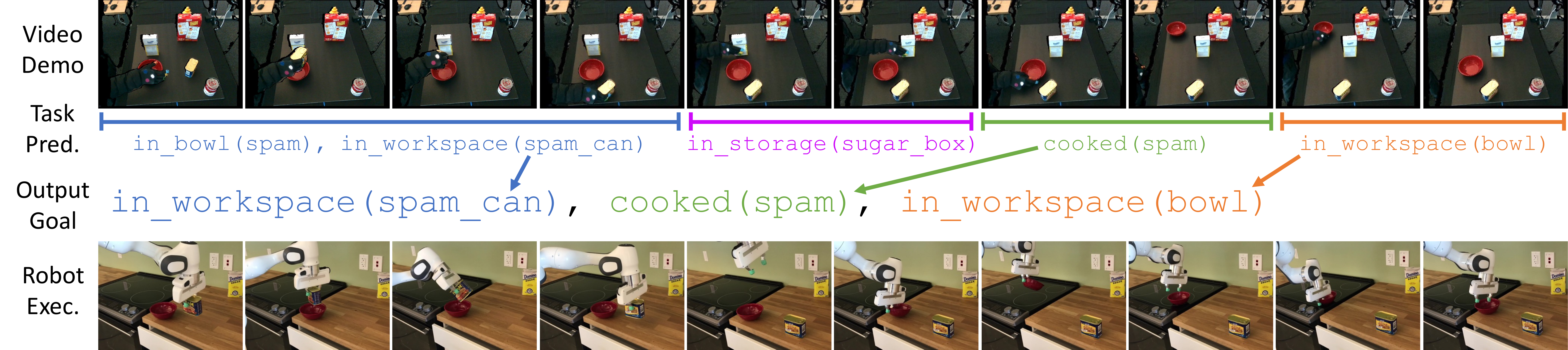}
  \figtop
  \caption{
    Qualitative results for third-person imitation from observation. Given the video demonstration at the top, our framework is able to successfully extract the intended goal. The goal is then used as input to a robot system for execution to reproduce the goal in a different environment.
  }
  \figbot
  \label{fig:qual_exec}
\end{figure*}

We are interested in problems that involve both task and motion reasoning. Cooking tasks are ideal because they involve multiple steps, and we have to potentially rearrange the objects to create valid paths. In addition to the standard pick-and-place operators/predicates, we introduce two additional operators and their associated predicates:
\begin{enumerate}
    \item \pred{pour(X,Y)}: \pred{in\_hand(X)}, \pred{on(X,Y)} $\Rightarrow$ \pred{in(X,Y)}
    \item \pred{cook(X)}: \pred{in(X,b)}, \pred{on\_stove(b)} $\Rightarrow$ \pred{cooked(X)}
\end{enumerate}
A further constraint for \pred{pour} is that $Y$ needs to be a container. We only use a bowl as the container. The kitchen environment is divided into four regions: workspace, storage, and two stoves. \figref{setup} shows the regions in both the mockup and the real kitchen.

We design our tasks as follow: a task is to cook an ingredient $F$, which involves 2 key steps \pred{pour} and \pred{cook}. One of the key steps would be initially blocked by a blocking object $B$, and the goal might or might not involve having $B$ to a new target region. We consider two ingredients $F=\{\text{tomato soup}, \text{spam}\}$ and three blocking objects $B=\{\text{cracker box}, \text{mustard bottle}, \text{sugar box}\}$. This gives a total of $2\times2\times3\times2=24$ tasks. We collect 4 demonstrations for each task. This results in a total of 96 demonstrations. We exclude videos with substantial missing poses from the evaluation for a meaningful comparison. An example demonstration is shown in \figref{demo}.

\subsection{Evaluating Goal Recognition}

\noindent\textbf{Experimental Setup.} We perform 10 random splits of the 24 tasks in our dataset: 12 for training and 12 for testing. For methods that do not require training (including ours), the training set is used to select the hyperparamters. 

\vspace{1mm}
\noindent\textbf{Baselines.} We compare to the following methods:

\noindent\textit{- Final State.} The Final State baseline just uses the true predicates in the final frame as goal~\cite{huang2019continuous}.

\noindent\textit{- Task Predicates.} The Task Predicates baseline uses all the achieved task predicates in the demonstration without analyzing the segments at the motion level. We apply the same predicate pooling scheme as ours for a fair comparison.

\noindent\textit{- Recurrent Neural Networks (RNN).} We use RNN as an end-to-end learning baseline. RNN uses the same object centric-representation $x_t$ as ours. The hidden states are averaged over time and classified with a two-layer MLP for the subgoals.

\noindent\textit{- Discrete Graphical Model (DGM)~\cite{chung2015bayesian}.} We compare to graphical model approaches for goal-recognition. The main difference is that in our case the action of a segment is uniquely defined by the start and end states, but the goal is not uniquely defined by the final state of the segment. We thus collect statistics for $P(G|X_i,X_f)$ instead of $P(A|X_i, X_f)$, where $X_i$ and $X_f$ are the start and end states of an action $A$ to achieve goal $G$.

\noindent\textit{- Ours w/o Motion ($m_i$).} We analyze the importance of our motion reasoning by comparing to an ablation without the motion reasoning. Instead of comparing $g=\bar{g}_i$ or $m_i(b_i)$ using \eqnref{goal_recog}, this baseline just looks at $g=\bar{g}_i$ and select a threshold for \eqnref{goal_recog} to see if the task predicate $\bar{g}_i$ is actually part of the goal.

\vspace{1mm}
\noindent\textbf{Metrics.} We consider the following metrics:

\noindent\textit{- F1 Score.} As the goal is a subset of all predicates, standard metrics are the Precision, Recall, and F1 Score to see how well the predicted goal match with the groundtruth.

\noindent\textit{- F1$_{blk}$.} As there are still many predicates in the goal, we further focus the evaluation on predicates of the blocking object ($blk$) because the blocking object is the main source of ambiguous subgoals in the mockup cooking domain.

\noindent\textit{- Success Rate.} We approximate the actual execution success rate by just looking at the output goal. If the output goal is exactly the same as the groundtruth goal, we treat it as a success and a failure otherwise.

\begin{table}[t]
\caption{Goal recognition results on our Mockup Cooking Dataset. }
 \label{tab:results}
 \centering
 \vspace{-3mm}
\begin{tabular}{l|cc|cccc}
\cline{1-6}
               & Prec.             & Recall             & F1            & F1$_{blk}$ & Succ.    &  \\ \cline{1-6}
Final State             & 0.31          & \textbf{0.98} & 0.47          & 0.53          & 0.0           &  \\
Task Pred.             & 0.74          & 0.96          & 0.81          & 0.57          & 0.24          &  \\ \cline{1-6}
RNN            & 0.72          & 0.56          & 0.62          & 0.47          & 0.17          &  \\
DGM~\cite{chung2015bayesian}            & 0.76          & 0.94          & 0.82          & 0.54          & 0.26          &  \\ \cline{1-6}
Ours w/o $m_i$ & \textbf{0.90} & 0.84          & 0.84          & 0.35          & 0.29          &  \\
Ours           & 0.88          & 0.91          & \textbf{0.88} & \textbf{0.76} & \textbf{0.47} & \\ \cline{1-6}
\end{tabular}
\vspace{-3mm}
\end{table}

\vspace{1mm}
\noindent\textbf{Results.} The results are shown in \tabref{results}. The Final State baseline has the highest recall because the goal predicates should be true in the last frame. However, the high recall is at the price of low precision. The Task Predicates baseline achieves much higher precision by recognizing all the high-level actions in the demonstration. We consider two learning based baselines: RNN and DGM~\cite{chung2015bayesian}. With less than 50 training videos, the learning based methods do not generalize well. DGM improves slightly over the Task Predicates by accumulating statistics of how a predicate is part of the goal. Without motion reasoning, none of the above baseline can handle the moving of the blocking objects and thus do not have reasonable F1$_{blk}$. Our full model explicitly performs motion reasoning about the objects in the demonstration, and thus would not blindly take all the object movements as intentional. This gives much higher precision compared to the Task Predicates baseline. Our approach thus best balance the precision and recall, and significantly outperforms the baselines ($+21\%$ for Succ. and $+19\%$ for F1$_{blk}$). We further analyze the importance of our motion reasoning. Ours w/o $m_i$ does not consider the motion predicate, but aims to recognize the goal by comparing how optimal the trajectory is to achieve the task predicate. This conservative baseline gives the highest precision, but at the cost of lower recall. In addition, the baseline is unable to handle the blocking object to complete the task without motion reasoning.

\subsection{Third-person Imitation from Observation}

One additional advantage of our goal-based framework is that it enables imitation learning across different environments. To demonstrate this, we executed on a Franka Panda robot in a real kitchen environment.
We used PoseCNN~\cite{xiang2017posecnn} to provide initial estimates as to object positions, and used DART~\cite{schmidt2014dart} to track objects as they moved around and to perform hand-eye calibration between the Franka robot and the kitchen. Primitive motion policies are executed via RMPflow~\cite{cheng2018rmpflow}, which allows for reactive, closed-loop control. We use the output goal from the video demonstration for task planning. The task plan is then
represented as a Robust Logical-Dynamical System~\cite{paxton2019representing} for reactive recovery and robustness to sensor noise in execution. 

\figref{qual_exec} shows qualitative results. Based on the video demonstration collected in the mockup kitchen, we are able to successfully extract the correct goal despite the manipulation of the sugar box. This extracted goal is input to the task planner from~\cite{paxton2019representing} and the resulting plan is successfully executed in the real kitchen. Although the demonstrator moves the sugar box in the video demonstration, and the sugar box also appears in the kitchen, the robot recognizes that it does not need to move the sugar box because it is already out of the way and the goal is just to cook the spam.

\section{Conclusion}

We present a new motion reasoning framework to recognize the goals from real-world video demonstrations. We show that despite being ambiguous at the symbolic action level and in terms of final states, the demonstrations are still intent-expressive at the trajectory level. By explicitly reasoning about object trajectories for task goals, we combine task and motion reasoning to infer the goal of the demonstration. Our results show that this allows us to significantly outperform previous approaches that aim to infer the goal based on either just motion planning or task planning. In addition, we show that our goal-based formulation enables the robot to reproduce the same goal in a real kitchen by just watching the video demonstration from a mockup kitchen.

\vspace{2mm}
\noindent\textbf{Acknowledgements.} We thank Ankur Handa, Yu Xiang, and Clemens Eppner for their help and discussions on the project.

%%%%%%%%%%%%%%%%%%%%%%%%%%%%%%%%%%%%%%%%%%%%%%%%%%%%%%%%
\renewcommand*{\bibfont}{\footnotesize}
\begin{flushright}
\printbibliography %for biblatex
\end{flushright}

\end{document}